\documentclass[10pt,twocolumn,letterpaper]{article}

\usepackage[pagenumbers]{cvpr} 

\usepackage[dvipsnames]{xcolor}

\usepackage{graphicx}
\usepackage{amsmath}
\usepackage{amssymb}
\usepackage{booktabs}
\usepackage{float}

\usepackage[accsupp]{axessibility}

\usepackage[dvipsnames]{xcolor}
\usepackage{todonotes}
\usepackage{bm}
\usepackage{multirow}
\usepackage{caption}
\usepackage{subcaption}
\usepackage{adjustbox}
	
\usepackage{color, colortbl}
\definecolor{Gray}{gray}{0.8}

\usepackage{mathtools}
\DeclarePairedDelimiter{\ceil}{\lceil}{\rceil}
\DeclarePairedDelimiter{\floor}{\lfloor}{\rfloor}

\usepackage{array}

\usepackage{makecell}
\usepackage{etoolbox}
\usepackage{caption}
\definecolor{cvprblue}{rgb}{0.21,0.49,0.74}
\usepackage[pagebackref,breaklinks,colorlinks,citecolor=cvprblue]{hyperref}

\def\paperID{13} 
\def\confName{CVPR}
\def\confYear{2024}

\title{FlowIBR: Leveraging Pre-Training for \\ Efficient Neural Image-Based Rendering of Dynamic Scenes}

\author{
Marcel B\"usching\textsuperscript{1} 
\and 
Josef Bengtson\textsuperscript{2} 
\and
David Nilsson\textsuperscript{2} 
\and
M{\aa}rten Bj\"orkman\textsuperscript{1} 
\and
\textsuperscript{1}KTH Royal Insitute of Technology
\and
\textsuperscript{2}Chalmers University of Technology
}

\begin{document}
\maketitle

\begin{abstract}
We introduce FlowIBR, a novel approach for efficient monocular novel view synthesis of dynamic scenes.
Existing techniques already show impressive rendering quality but tend to focus on optimization within a single scene without leveraging prior knowledge, resulting in long optimization times per scene.
FlowIBR circumvents this limitation by integrating a neural image-based rendering method, pre-trained on a large corpus of widely available static scenes, with a per-scene optimized scene flow field.
Utilizing this flow field, we bend the camera rays to counteract the scene dynamics, thereby presenting the dynamic scene as if it were static to the rendering network.
The proposed method reduces per-scene optimization time by an order of magnitude, achieving comparable rendering quality to existing methods — all on a single consumer-grade GPU.
\end{abstract}    
\section{Introduction}
\label{sec:intro}

\begin{figure}
    \centering
    \includegraphics[width=0.98\columnwidth]{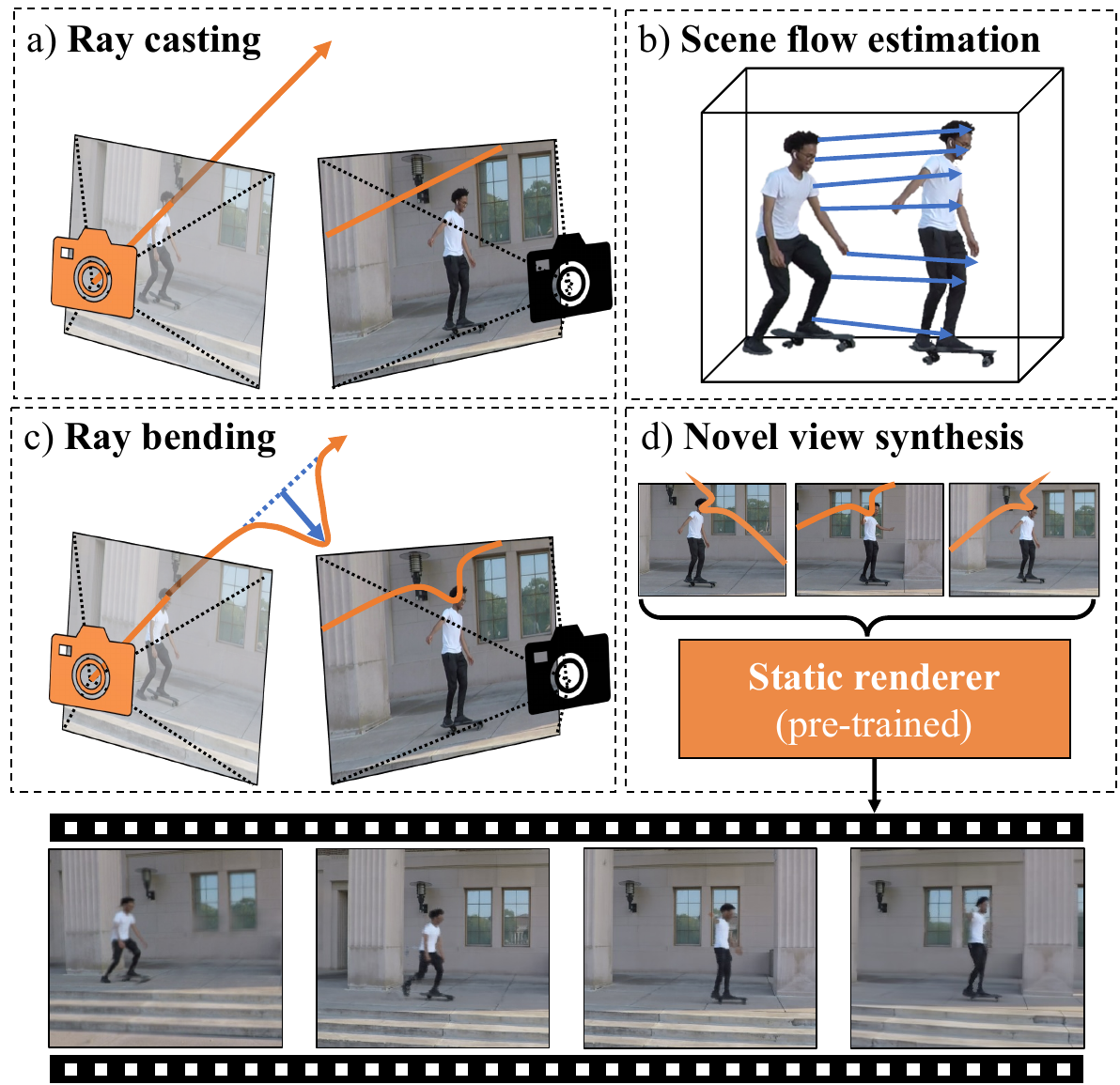}
    \caption{
    \textbf{Method overview}
    a) An image at an arbitrary position (orange camera) is synthesised based on existing observations (black camera), collected at different times. Problem: Due to the movement of the skater, the skater is not on the epipolar line of the camera ray.
    b) We model scene motion using per-scene learned scene flow.
    c) Scene flow is used to compensate the motion by bending the camera ray.
    d) A pre-trained neural IBR method for static scenes \cite{t2023AttentionAll} is used for image synthesis.} 
\label{fig:method}
\end{figure}

Novel view synthesis for dynamic scenes allows for rendering views of an observed scene from new viewpoints, possibly also at new points in time. 
Recent methods \cite{park2021NerfiesDeformable, pumarola2021DNeRFNeural, li2021NeuralScene, li2023DynIBaRNeural} already show impressive rendering qualities.
Nevertheless, they suffer from long training times and show limitations for fast-changing scenes with sparse observations \cite{gao2022MonocularDynamic}. We assume that these limitations are partially due to the fact that these methods are only optimized per scene without exploiting any prior knowledge.
Therefore, we present FlowIBR which combines a per-scene learned scene flow field with a pre-trained generalizable novel view synthesis method \cite{t2023AttentionAll} as a rendering backbone.

For static scenes there is already a multitude of generalizable novel view synthesis methods \cite{wang2021IBRNetLearning, t2023AttentionAll, suhail2022GeneralizablePatchBased, liu2022NeuralRays, sajjadi2022SceneRepresentation} which are trained on many different scenes to learn how to aggregate information from observations to synthesize novel views.
Importantly, generalizable methods can therefore render novel views of previously unseen scenes \emph{without} any scene-specific training.
These methods often outperform per-scene optimized methods in terms of rendering quality, training times and demand for dense observation of the scene \cite{wang2021IBRNetLearning, liu2022NeuralRays, yu2021PixelNeRFNeural}.
However, due to the limited availability of datasets of dynamic scenes for training, such generalized methods cannot readily be extended to dynamic scenes.

To overcome this limitation, we utilize a pre-trained rendering backbone based on Generalizable NeRF Transformer (GNT) \cite{t2023AttentionAll}, a generalizable view synthesis method which has been pre-trained on a large corpus of more easily obtainable static scenes.
GNT performs neural image-based rendering (IBR) \cite{wang2021IBRNetLearning}, synthesizing a novel view by pixel-wise  projecting camera rays as epipolar lines into neighbouring source views and then aggregating image information along them.
However, in dynamic scenes, projections of independently moving scene content are likely displaced with respect to the epipolar lines.
Therefore, we use a per-scene learned scene flow field to bend camera rays, so they follow the motion of the scene content over time as shown in \cref{fig:method}.
The core optimization is thus simplified by shifting focus from jointly learning the complex interplay of scene geometry, color, and temporal changes to a separate step of learning the scene dynamics.

With this approach, we are able to reduce the necessary per-scene setup time to about 1.5 hours, an improvement by an order of magnitude compared to previous methods, while getting comparable rendering quality.
Additionally, this dynamics-focused optimization process allows us to create a training regime which enables optimizing the scene flow network on a single Nvidia 3080 RTX GPU.
In summary, our contributions are:
\begin{itemize}
    \item \textit{Introducing FlowIBR}, a novel view synthesis method for dynamic scenes which combines a per-scene optimized scene flow field with a  pre-trained neural image-based rendering method to decrease training time.
    \item \textit{Presenting a dynamics-centred training regime} which allows for fast training of the proposed method on a single consumer-grade GPU.
    \item \textit{Demonstrating the performance of FlowIBR} on the Nvidia Dynamic Scenes Dataset \cite{yoon2020NovelView} showing competitive rendering quality despite significantly lower training times with respect to the state-of-the-art.
\end{itemize}
Code for the proposed method will be made available\footnote{\url{https://github.com/buesma/flowibr}}.
\section{Related Work}
\label{sec:related-work}

\paragraph{Novel view synthesis for static scenes:}
Neural Radiance Fields (NeRFs) \cite{mildenhall2020NeRFRepresenting} and succeeding methods \cite{barron2021MipNeRFMultiscale,barron2022MipNeRF360, zhang2020NeRFAnalyzing, muller2022InstantNeural} are able to synthesize photo-realistic images of a scene by modelling a continuous radiance and density function of an observed scene with a multi-layer perceptron (MLP).
Gaussian Splatting \cite{kerbl20233DGaussian} represents the scenes with a set of 3D Gaussians, which are projected on the image plane and then aggregated to a pixel color via GPU optimized alpha blending.
NeRF and Gaussian Splatting based methods are optimized per scene without leveraging prior knowledge.
In contrast, techniques for generalizable novel view synthesis are trained across multiple scenes, in order to allow for synthesis of novel views from sparsely observed scenes, unseen during training.
Earlier methods such as pixelNeRF \cite{yu2021PixelNeRFNeural} and MVSNeRF \cite{chen2021MVSNeRFFast} have achieved this by deploying a generalizable NeRF conditioned on latent vectors extracted from the source observations.
Many of the current methods \cite{wang2021IBRNetLearning, liu2022NeuralRays, suhail2022LightField, suhail2022GeneralizablePatchBased, reizenstein2021CommonObjects, t2023AttentionAll, du2023LearningRenderNovel, johari2022GeoNeRFGeneralizing} combine transformers \cite{dosovitskiy2022ImageWorth, vaswani2017AttentionAll} with multi-view geometry.
As proposed by IBRNet \cite{wang2021IBRNetLearning}, this is realized by projecting points sampled along the camera ray for each pixel into the source observations. 
The information at these projected points is then extracted and subsequently aggregated into pixel color using transformers — often in combination with volumetric rendering.
GNT \cite{t2023AttentionAll} and a recent method from Du~\etal~\cite{du2023LearningRenderNovel} are fully transformer-based approaches, replacing the volumetric rendering, with a learned rendering.


\paragraph{Novel view synthesis for dynamic scenes:}


While early techniques required multi-view videos \cite{lombardi2019NeuralVolumes, li2022Neural3D, wang2022FourierPlenOctrees}, current NeRF-based methods have demonstrated remarkable rendering capabilities from monocular videos. 
Methods such as D-NeRF \cite{pumarola2021DNeRFNeural} or Nerfies \cite{park2021NerfiesDeformable} extend NeRFs to the dynamic domain by learning a canonical NeRF representation at a fixed point in time \cite{pumarola2021DNeRFNeural, park2021NerfiesDeformable, park2021HyperNeRFHigherdimensional, guo2023ForwardFlow, song2023NeRFPlayerStreamable}, which is then warped by a learned motion field to other instances in time.
This canonical scene representation serves as an anchor in time, accumulating the geometric and color information from source observations.
A different line  of work \cite{kundu2022PanopticNeural, ost2021NeuralScene} instantiates moving objects as separate bounded NeRFs inside a static environment NeRF, which are then rigidly moved between time-steps by learned transformations.
For non-rigid dynamics, several methods \cite{ xian2021SpacetimeNeural, li2021NeuralScene, gao2021DynamicView} learn time-varying NeRF representations of the scene, often decomposing the scene in static and dynamic content \cite{gao2021DynamicView, li2021NeuralScene, zhang2023DetachableNovel}.
To overcome the ambiguity in observing non-rigid dynamic scenes with a single camera, methods such as NSFF \cite{li2021NeuralScene}, DVS \cite{gao2021DynamicView}, NeRFlow \cite{du2021NeuralRadiance} and DFNet \cite{yoon2020NovelView} also estimate the scene flow and use it in combination with optical flow \cite{teed2020RAFTRecurrent}, depth estimates \cite{ranftl2022RobustMonocular} and multi-view geometry as additional supervision for the per-scene learned dynamic NeRF.
DVS also explores the concept of splitting the learning process — initially pre-training a static rendering framework and subsequently employing it to simplify the training of a dynamic approach. 
However, DVS optimizes both components only for a single scene.
A recent method DynIBaR \cite{li2023DynIBaRNeural}, which is based on NSFF, also utilizes neural IBR for novel view synthesis with techniques to address scene motion.
The key distinction lies in DynIBaR being fully optimized on a single scene without utilizing pre-training, leading to long training times of up to 2 days on 8 GPUs.

\paragraph{Efficient view synthesis}
The aforementioned methods are optimized for improved visual rendering quality, but come with long per-scene training times.
This problem is the object of several recent methods which introduce efficient data structures \cite{fridovich-keil2023KPlanesExplicit, cao2023HexPlaneFast, wang2023MaskedSpaceTime, wang2023MixedNeural, shao2023Tensor4DEfficient, guo2023ForwardFlow, fang2022FastDynamic, song2023NeRFPlayerStreamable} or utilize Gaussian Splatting \cite{wu20234DGaussian, luiten2024Dynamic3D, duisterhof2023MDSplattingLearning} to reduce the computational complexity and convergence time.
NeRFPlayer \cite{song2023NeRFPlayerStreamable}, MSTH \cite{wang2023MaskedSpaceTime} and MixVoxels \cite{wang2023MixedNeural} simplify the training process further by decomposing the scene in static and dynamic content.
Although those methods show substantial improvements in rendering and training times, they rely on multi-view images as in the Plenoptic Video dataset \cite{li2022Neural3D} or quasi multi-view images \cite{gao2022MonocularDynamic} as in the D-NeRF dataset  \cite{pumarola2021DNeRFNeural} or HyperNeRF dataset \cite{park2021HyperNeRFHigherdimensional}.
We base our method on easier to obtain monocular images, and evaluate on, according to the survey from Gao~\etal \cite{gao2022MonocularDynamic}, more complex Nvidia Dynamic Scenes Dataset \cite{yoon2020NovelView, li2021NeuralScene}.
%


Similar to our approach, Fourier PlenOctree \cite{wang2022FourierPlenOctrees} and MonoNeRF \cite{tian2023MonoNeRFLearning} utilize generalizable novel view synthesis for dynamic scenes.
However, PlenOctree also requires multi-view images that are harder to obtain than monocular videos.
MonoNeRF pursues the goal of learning a fully generalizable dynamic radiance field by utilizing temporal features extracted from videos to estimate the scene flow which is then used to aggregate spatial features to pixel colors. Nevertheless, for the generalizable setting where MonoNeRF is pre-trained on one scene and then fine-tuned on the remaining scenes, only LPIPS values are reported. These are several times higher than those of the per-scene setting, for which the performance is more competitive, but at the cost of 4h of training time per scene on an A100 GPU.

\section{Method}
\label{sec:method}

\paragraph{Problem formulation:}
Given a set of images $\bm{I}_t \in \mathbb{R}^{H\times W \times 3}$ taken at discrete times $t \in \{1, \ldots, T\}$ with known camera matrices $\bm{P}_t \in \mathbb{R}^{3 \times 4}$, we denote the triplet $(\bm{I}_t, \bm{P}_t, t)$  as an observation of the scene.
We assume observations to be taken in constant time intervals $\Delta t$.
The problem is now to synthesize a target view of the scene $\Tilde{\bm{I}}$ from an \textit{arbitrary} viewpoint defined by $\Tilde{\bm{P}}$ and \textit{continuous} time $\Tilde{t} \in \mathbb{R}$, based on the set of available observations
\begin{equation}
    O = \{(\bm{I}_t, \bm{P}_t, t)\}_{t=1}^T.
\end{equation}

\paragraph{Overview:} 
Our proposed method combines a pre-trained generalizable novel view synthesis method with a scene flow field which is learned per scene.
In this section, we will review the rendering backbone used for static scenes, introduce the scene flow field model, describe the loss and regularization terms, and propose a training regime for fast training on a single GPU.

\subsection{Pre-training on static scenes}
\label{subsec:method}


For our rendering backbone we build upon Generalizable NeRF Transformer (GNT) \cite{t2023AttentionAll},  a transformer-based extension of IBRNet\cite{wang2021IBRNetLearning}.
GNT is able to synthesize novel views of a \textit{static} scene not seen during training, at an arbitrary target viewpoint.
Novel view synthesis is performed pixel-wise in a two-stage process that involves two transformers.

\textit{View Transformer (VT)}: 
This initial stage encompasses aggregating information from source observations close to the target view.
This is achieved by firstly encoding each source observation into a feature map using a U-Net \cite{ronneberger2015UNetConvolutional} image encoder $\bm{F}_t = \text{U-Net}(\bm{I}_t)$.
Using the camera matrix of the target view, a camera ray $\bm{r}_{t,m}$ is cast from the focal point $\bm{e}_t$ of the camera  through the target pixel $m \in \{1, \ldots, H \times W\}$ in ray direction $\bm{d}_{t,m}$,
\begin{equation}
    \bm{r}_{t,m}(l) = \bm{e}_t + l \bm{d}_{t,m}.
\end{equation}
Afterwards, $N$ points $\bm{p}_{t,m,n} = \bm{r}_{t,m}(l_n)$ are sampled at different distances $l_n\in\mathbb{R}$ along this ray and projected onto each considered source observation.
To simplify notation, we omit $n$ and $m$ indexes wherever possible.
Consequently, these projected points are located on the respective epipolar lines, defined by the target pixel and the image planes of the source observations.
Image features are extracted by interpolating the feature map at these projected points. 
For every individual point along the camera ray, the view transformer aggregates the associated image feature vectors through attention into a single feature vector, representing the scene content at that particular spatial location.

\textit{Ray Transformer (RT):} 
This stage utilizes a ray transformer to conduct a learned ray-based rendering using the feature vectors created by the view transformer in the first stage.
Using attention, the features along the ray are aggregated into a unified feature representation of that ray.
A final multi-layer perceptron (MLP) is then applied to decode this ray feature vector, translating it into the corresponding RGB value for that particular ray.


\paragraph{Optimization:} 
GNT is trained by using one of the observations as target, while employing the remaining observations as source observations.
This is a technique commonly found in NeRFs \cite{mildenhall2020NeRFRepresenting} where a color prediction loss term
\begin{equation}
    \mathcal{L}_{rgb} = \sum_{m} || \bm{C}_{m} - \hat{\bm{C}}_{m} ||^2_2,
\end{equation}
 is applied to the final pixel color $\bm{C}_{m}$ and ground truth $\hat{\bm{C}}_{m}$ color of pixel $m$.
In contrast to NeRF, GNT is trained for each training step on a different scene.
This enables novel view synthesis for scenes not observed during training, without additional optimization.
We select GNT as rendering backbone for our work, based on its state-of-the-art generalization capabilities and rendering quality.
For more details about GNT, we refer to the original paper \cite{t2023AttentionAll}.

\begin{figure}
    \centering
    \includegraphics[width=0.85\columnwidth]{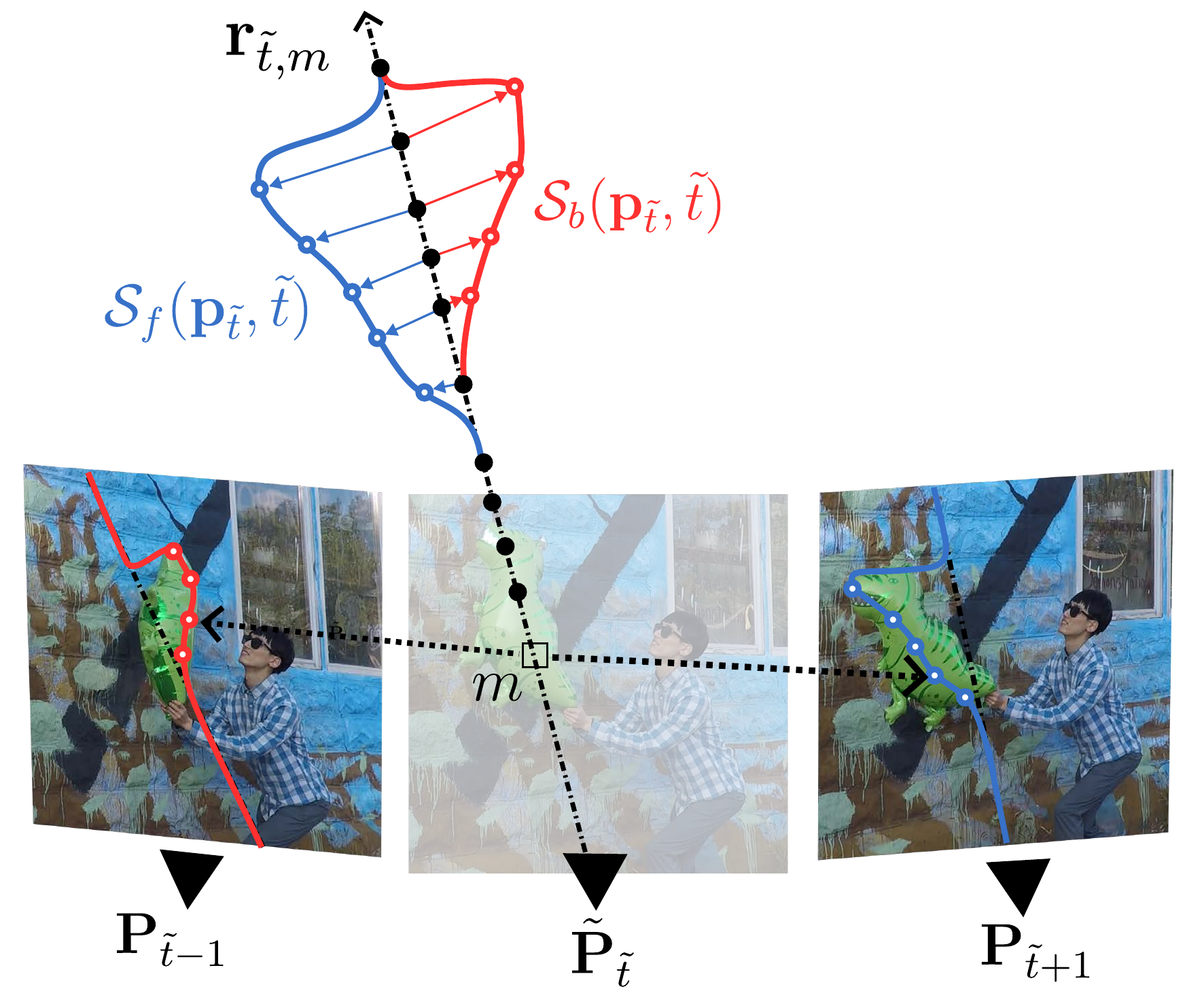}
    \caption{\textbf{Scene flow compensation}
    The scene flow ($\mathcal{S}_{f}$, $\mathcal{S}_{b}$) is used to adjust the ray $\bm{r}_{\Tilde{t},m}$ from the target camera $\Tilde{\bm{P}}_{\Tilde{t}}$ through the current pixel $m$, so that it follows the motion of the balloon at the two adjacent times 
    $\Tilde{t}+1$ and $\Tilde{t}-1$.
    This allows the projection of the ray on the source observations to contain the pixels corresponding to $m$, marked by arrows.}
    \label{fig:scene-flow-compensation}
\end{figure}

\subsection{Scene flow field}

Naively applying neural image-based rendering to dynamic scenes will yield insufficient results (see \cref{subsec:ablation}) due to the assumption that relevant neural image features can be aggregated along the epipolar lines of the target pixel on the observations.
However, in dynamic scenes where each observation captures a distinct state, this faces challenges due to potential misalignment between the potentially moving scene content and the epipolar lines.
Our goal is therefore to learn the underlying scene motion as scene flow fields \cite{li2021NeuralScene, gao2021DynamicView} in order to be able to realign the neural feature aggregation with the dynamic image content.

We represent per-scene learned scene flow as a forward flow field
\begin{equation}
     \mathcal{S}_{f}: (\bm{p}_t, t) \rightarrow \bm{s}_{f}
\end{equation}
which maps each time $t \in \mathbb{R}$ and position $\bm{p}_t\in \mathbb{R}^3$ to a forward scene flow vector $\bm{s}_{f} \in \mathbb{R}^3$ and a backward flow field
\begin{equation}
     \mathcal{S}_{b}: (\bm{p}_t, t) \rightarrow \bm{s}_{b}
\end{equation}
 which maps to a backward scene flow vector $\bm{s}_b \in \mathbb{R}^3$.
These scene flow vectors can be used to displace a point from one time to an adjacent time step with
\begin{align}
    \bm{p}_{t\rightarrow t+1} &= \bm{p}_{t} + \mathcal{S}_{f}(\bm{p}_{t}, t) \\
    \bm{p}_{t\rightarrow t-1} &= \bm{p}_{t} + \mathcal{S}_{b}(\bm{p}_{t}, t)
\end{align}
where we generally denote $\bm{p}_{t\rightarrow t+1}$ as the position $\bm{p}_{t}$ adjusted to the next time $t+1$ and $\bm{p}_{t\rightarrow t-1}$ to the previous time $t-1$ .
The fields are represented by two different heads of the same MLP
\begin{equation}
    \text{MLP}_{\bm{\theta}}: (\psi(\bm{p}), t) \rightarrow (\bm{s}_{f}, \bm{s}_{b}).
\end{equation}
To decrease the computational complexity of estimating the scene flow, we combine a shallow MLP with the multi-resolution permutohedral lattice from PermutoSDF \cite{rosu2023PermutoSDFFast}.
The number of vertices in common voxel \cite{liu2020NeuralSparse} or Instant-NGP \cite{muller2022InstantNeural} based data structures grows exponentially with the input dimension, which quickly leads to a poor memory footprint and convergence time.
Differently, permutohedral lattices scale linearly with the input dimension and have been shown to work well with the 4D inputs used in our case \cite{rosu2023PermutoSDFFast}. 
Details of the scene flow network architecture are illustrated in the supplementary materials.

\subsection{Combined method}

As introduced in \cref{fig:scene-flow-compensation}, we utilize the scene flow to adjust the 3D position of the points along the camera ray.
This essentially \textit{bends} the camera ray \cite{pumarola2021DNeRFNeural} — ensuring the ray points are projected to the parts of the image plane that contain dynamic content.
Therefore, the camera rays will follow the moving scene content and thus allow static image-based rendering from being affected by scene motion, which allows for its use in dynamic scenes without any adjustments to the basic formulation.

For using more than just the two images closest to the target time $\Tilde{t}$ as source observations, it is necessary to estimate the scene flow $\bm{s}_{\Tilde{t}\rightarrow t}$ from the target over a larger time window.
This can be done with an iterative function evaluation of the scene flow network.
\begin{equation}
    \mathcal{S}_{\Tilde{t} \rightarrow t}(\bm{p}_{\Tilde{t}}) 
        = \begin{cases}
          \mathcal{S}_f (\bm{p}_{\Tilde{t}}, \Tilde{t})
           + \mathcal{S}_{\Tilde{t}+1\rightarrow t}(\bm{p}_{\Tilde{t}+1\rightarrow t})  
& , \Tilde{t}<t \\
          \mathcal{S}_b (\bm{p}_{\Tilde{t}}, \Tilde{t})
           + \mathcal{S}_{\Tilde{t}-1\rightarrow t} (\bm{p}_{\Tilde{t}-1\rightarrow t})
& , \Tilde{t}>t \\
     \bm{0} & , \Tilde{t}=t
    \end{cases}
    \label{eq:multistep-scene-flow}
\end{equation}


This stays computationally feasible when selecting a sufficiently small MLP (e.g. 6  to 8 layers) for the scene flow, and basing the image synthesis process on images which are taken at temporally close to the target time $\mathcal{N}(\Tilde{t}) \ni t$, to limit the number of times to adjust the rays for. 

Rendering at continuous target times $\Tilde{t} \in \mathbb{R}$ outside the intervals $\Delta t$, in which the scene has been observed, is facilitated by estimating the scene flow at the target time, and then linearly scaling the scene flow vectors, so they displace the points not over the full interval $\Delta t$ but to the times of the next observations. 
Following this initial step, the motion adjustment can continue as previously described.
A more formal introduction of this can be found in the supplementary materials.


\subsection{Losses}

The problem of learning a non-rigid dynamic scene from monocular observations is highly ambiguous. 
For example, an object that has seemingly grown from one observation to the next can either have changed its actual size or just moved towards the camera.
This ambiguity creates the need for additional supervision and regularization besides the previously introduced $\mathcal{L}_{rgb}$ loss.
For the selection of losses that are used, we take inspiration from previous scene flow-based works \cite{li2021NeuralScene, li2023DynIBaRNeural, gao2021DynamicView}.

\paragraph{Optical flow loss:}
Given the absence of a ground truth for the scene flow, the optical flow between the target image and source observations serves as an effective proxy which can be used for additional supervision.
For this we use RAFT \cite{teed2020RAFTRecurrent} to estimate the optical flow $\bm{o}_{\Tilde{t}\rightarrow t}$ from the target image at time $\Tilde{t}$ to the used source observations at times $t$.

The optical flow estimate for each pixel is a 2D vector that represents the displacement to the corresponding pixel location in a second image.
We use this as supervision for the 3D scene flow, by taking the scene flow-adjusted points along the target ray, and projecting them onto the image plane of the observations. 
Then we calculate the difference between the pixel locations of the projected points and the pixel in the target image corresponding to the camera ray.
This yields one pixel displacement vector per projected ray point.
We use the attention weights of the ray transformer to estimate a weighted average $\bm{d}_{\Tilde{t}\rightarrow t}$ of these. 
The optical flow loss is then calculated with the L1-norm between the optical flow $\bm{o}_{\Tilde{t}\rightarrow t}$ estimated with RAFT and the weighted average $\bm{d}_{\Tilde{t}\rightarrow t}$ of projected scene flow,
\begin{equation}
    \mathcal{L}_{of} = \sum_{t\in\mathcal{N}(\Tilde{t})} ||\bm{o}_{\Tilde{t}\rightarrow t} - \bm{d}_{\Tilde{t}\rightarrow t} ||_1.
\end{equation}
This procedure is comparable to related methods \cite{li2021NeuralScene, li2023DynIBaRNeural} which weigh pixel displacements with the opacity values of the NeRF representation.
Due to noise in the optical flow, we only use $\mathcal{L}_{of}$ to initialize the scene flow to a general direction, and linearly anneal its weight during training.


\paragraph{Cycle consistency regularization:}
To ensure that the learned forward and backward scene flows are consistent to each other, we utilize \textit{cycle consistency} regularization \cite{gao2021DynamicView, li2021NeuralScene, li2023DynIBaRNeural}.
This means that the backwards scene flow of point $\bm{p}_{t}$ should be equal to the forward scene flow of the displaced point $\bm{p}_{t\rightarrow t-1}$, but in the opposite direction,
\begin{equation}
\begin{split}
    \mathcal{L}_{cyc} =  \sum_{t\in\mathcal{N}(\Tilde{t})}&  
    || \mathcal{S}_{b}(\bm{p}_{t}, t) + \mathcal{S}_{f}(\bm{p}_{t\rightarrow t-1}, t-1) ||_1  \\
 +& || \mathcal{S}_{b}(\bm{p}_{t\rightarrow t+1}, t+1) + \mathcal{S}_{f}(\bm{p}_{t}, t) ||_1.
\end{split}
\end{equation}

\paragraph{Scene flow regularisation:}
We introduce three supplementary regularization terms to the scene flow, guiding it to learn anticipated properties of the inherent scene motion. 
For this we use the squared L2-norm between forward and backward scene flow as \textit{temporal smoothness} regularization \cite{vo2016SpatiotemporalBundle, li2021NeuralScene, gao2021DynamicView, li2023DynIBaRNeural}
\begin{equation}
    \mathcal{L}_{temp} = \sum_{t\in\mathcal{N}(\Tilde{t})} || \mathcal{S}_{f}(\bm{p}_{t}, t) + \mathcal{S}_{b}(\bm{p}_{t}, t) ||^2_2 \label{eq:reg-temporal}
\end{equation}
which encourages the learning of a piece-wise linear solution. 
Additionally, we use the L1-norm on the predicted scene flow to support a generally \textit{slow} scene flow, based on the common assumption, that most of the scene does not contain motion \cite{li2021NeuralScene, gao2021DynamicView, valmadre2012GeneralTrajectorya}.
\begin{equation}
    \mathcal{L}_{slow} = \sum_{t\in\mathcal{N}(\Tilde{t})} || \mathcal{S}_{f}(\bm{p}_{t}, t) ||_1 + || \mathcal{S}_{b}(\bm{p}_{t}, t) ||_1    \label{eq:reg-slow}
\end{equation}

Lastly, we encourage \textit{spatial smoothness} so that narrow points along a ray are going to have similar scene~flow~\cite{newcombe2015DynamicFusionReconstruction, li2021NeuralScene,  li2023DynIBaRNeural}.
For this, the difference between the forward and backward flows of neighbouring points $\bm{p}'\in\mathcal{N}(\bm{p})$ is weighted based on their distance
\begin{equation}
    w(\bm{p}', \bm{p}) = \exp(-2||\bm{p}'-\bm{p}||^2_2).   \label{eq:reg-spatial-weight}
\end{equation}
The loss is then calculated on the L1-norm between the deviations in the flow.
\begin{equation}
    \begin{split}
    \mathcal{L}_{spat} &= \sum_{t\in\mathcal{N}(\Tilde{t})} \sum_{\bm{p}'\in\mathcal{N}(\bm{p}_{t})}  
    \hspace{-8pt} (  || \mathcal{S}_{f}(\bm{p}_{t}, t) - \textit{S}_{f}(\bm{p}', t)  ||_1 + \\
    & +  || \mathcal{S}_{b}(\bm{p}_{t}, t) - \mathcal{S}_{b}(\bm{p}', t)  ||_1 ) \times w(\bm{p}', \bm{p}_{t}) 
    \end{split}
\end{equation}
We summarize these losses to an aggregated regularization loss term
\begin{equation}
\mathcal{L}_{reg} =  \mathcal{L}_{temp} + \mathcal{L}_{slow} + \mathcal{L}_{spat}.
\end{equation}

\paragraph{Final loss:}
The final total loss is calculated by a weighted summation over the different loss terms
\begin{equation}
    \mathcal{L} = \mathcal{L}_{rgb} + \alpha^k_{of} \mathcal{L}_{of} + \alpha_{cyc} \mathcal{L}_{cyc} + \alpha_{reg}\mathcal{L}_{reg},
\end{equation}
with  $\alpha^k_{of}$ being linearly annealed over $k$ steps of training.

\subsection{Dynamics focused optimization}

In the following, we will describe details of the process which will allow for optimization on a single GPU.
For this we take advantage of the aspect that the optimization can focus on learning the scene flow, since the rendering has already been learned during pre-training.
The general idea is to follow a coarse-to-fine approach \cite{park2021NerfiesDeformable} in learning scene flow, focusing on eliciting a coarse scene flow at the beginning, and then refining it as training time progresses.

\paragraph{Low-to-high number of source images:}
Training starts with a low number of source images to initially learn the scene flow between close time steps.
Once the scene flow has started to emerge, we steadily increase the maximum number of source images at pre-defined steps to also learn more nuanced long-term dynamics.
To keep a constant maximum GPU utilization, 
we adjust the ray batch size so that the total number of rays for all views remains constant, with a larger batch size in the beginning and then smaller batch sizes as the number of views increases. 

\paragraph{Coarse-to-fine image resolutions:}

Directly training the model on the original image resolution often leads to a zero collapse of the scene flow, i.e.~the scene is incorrectly predicted to be fully static.
To avoid falling into a trivial local minimum created by high-frequency details, we apply a coarse-to-fine scheme with the images originally subsampled by a factor of $f=8$.
The subsampling reduces the overall complexity of the correspondence problem, and the batches of rays are more likely to cover also small dynamic areas.
During training, we incrementally increase the image resolution by reducing $f$ in steps of $2$ to gradually refine the scene flow to represent the finer details of the underlying scene dynamics. 

\paragraph{Masked ray sampling:}
The regularization factors in $\mathcal{L}_{reg}$ benefit learning of a near-zero scene flow in static regions of the scene.
This allows us to focus the computational resources on the dynamic regions of the scene. We do so using binary motion masks that represent a coarse segmentation of dynamic regions in the image.
Instead of sampling pixels uniformly across the image, we increase the sampling probability of pixels in regions that are potentially dynamic. 

Motion masks are estimated by using an off-the-shelf instance segmentation network \cite{he2017MaskRCNN}.
Of the detected instances, we retain potentially dynamic classes (e.g. person, car) with a confidence above an intentionally small  threshold ($p=0.1$) to avoid discarding dynamic content.
Overly small instances are removed via morphological opening, where we use a large dilation filter mask to create a wide boundary around the remaining segments which allows for learning a better transition from static to dynamic regions.
The segmentation mask also allows us to downscale the $\mathcal{L}_{slow}$ loss at potentially dynamic pixels, encouraging a stronger scene flow.
Li~\etal~\cite{li2021NeuralScene} also utilize dynamic masks, but use them in a separate stage at the beginning of the training to sample additional rays in dynamic regions.

\paragraph{Fine-tuning of rendering backbone:}
To further increase the rendering quality we also fine-tune the rendering backbone per-scene.
This is possible with minimum computational overhead, since GNT is part of the optimization loop of the scene flow network.
We found this especially beneficial in combination with the optical flow loss, since the fine-tuning leads to refined attention weights of the ray transformer and therefore a more refined optical flow loss.
\section{Experimental Results}
\label{sec:evaluation}

In this section, we show the capabilities of our method, compare it to state-of-the-art baselines and ablations in quantitative and qualitative experiments, as well as analyse the learned scene flow.

\begin{table*}[ht!]
    \centering
    \caption{
        \textbf{Quantitative evaluation on Nvidia Dynamic Scenes \cite{yoon2020NovelView}}
         We report metrics as average over \underline{all} 8 scenes for the whole image as well as dynamic parts only.
         For NSFF we report results on two different training checkpoints with the number of train steps in brackets. FlowIBR (ours) has a significantly lower training time than the compared methods while obtaining competitive rendering quality.}
    \label{tab:comparison}
    \small
    \begin{tabular}{l|l| crr|cr|w{c}{0.72cm} w{c}{0.72cm} w{c}{0.72cm}|w{c}{0.72cm} w{c}{0.72cm} w{c}{0.72cm}}
    \Xhline{4\arrayrulewidth}
    \multirow{ 2}{*}{Data} & \multirow{2}{*}{Method}     &   \multicolumn{3}{>{\columncolor[gray]{0.9}}c|}{Train Time} &   \multicolumn{2}{c|}{Render Time}    & \multicolumn{3}{c|}{Full Image}  & \multicolumn{3}{c}{Dynamic Regions}  \\
                                & & \cellcolor[gray]{0.9}$\text{N}_{\text{GPU}}$ & \cellcolor[gray]{0.9}h \; \; & \cellcolor[gray]{0.9}GPUh  & $\text{N}_{\text{GPU}}$ & GPUsec         & PSNR $\uparrow$ &  SSIM $\uparrow$ & LPIPS $\downarrow$  & PSNR $\uparrow$ &  SSIM $\uparrow$ & LPIPS $\downarrow$\\
    \Xhline{3\arrayrulewidth}
    \multirow{5}{*}{default} &
    HyperNeRF       & 4 & 16:00 & 64:00 & 4    &  10.0\; \;  & 20.03 & 0.481 & 0.212 & 17.40 & 0.327 & 0.301\\
    & DVS           & 4 & 18:30 & 74:00 & \textbf{1} & 20.4\; \;        & 25.98 & 0.730 & 0.083 & \textbf{22.12} & \textbf{0.690} & 0.152 \\
     & NSFF (30k)   & 4 & 2:05 & 8:20 & \textbf{1}  & \textbf{6.2}\; \;  & 26.57 & 0.604 & 0.103 & 20.84 & 0.627 & 0.178 \\
    & NSFF (1M)     & 4 & 58:15 & 233:00 & \textbf{1}   & \textbf{6.2}\; \;  & \textbf{28.20} & 0.681 & \textbf{0.044} & 21.75 & 0.681 & \textbf{0.103} \\

    \cline{2-13} 
    & FlowIBR (Ours) & \textbf{1} & \textbf{1:27} & \textbf{1:27} & \textbf{1} &  11.8 \; &  26.81 & \textbf{0.742} & 0.099 & 21.13 & 0.651 & 0.192  \\
    \Xhline{3\arrayrulewidth}
    \multirow{2}{*}{long} &
    DynIBaR$^\dag$ & 8 & 48:00 & 384:00 & \textbf{1} &  20.0\; \;   & \textbf{30.92} & \textbf{0.958} & \textbf{0.027} & \textbf{24.32} & \textbf{0.827} & \textbf{0.061}\\
    \cline{2-13} 
    & FlowIBR (Ours) & \textbf{1} & \textbf{1:32} & \textbf{1:32} & \textbf{1} &  \textbf{11.3} \; & 27.32 & 0.793 & 0.083 & 22.06 & 0.702 & 0.132  \\
    \Xhline{3\arrayrulewidth}
    \multicolumn{13}{l}{\footnotesize{$\dag$ Results were taken from the original publication}}
    \end{tabular}

\end{table*}

\begin{figure*}[t]
    \centering
    \input{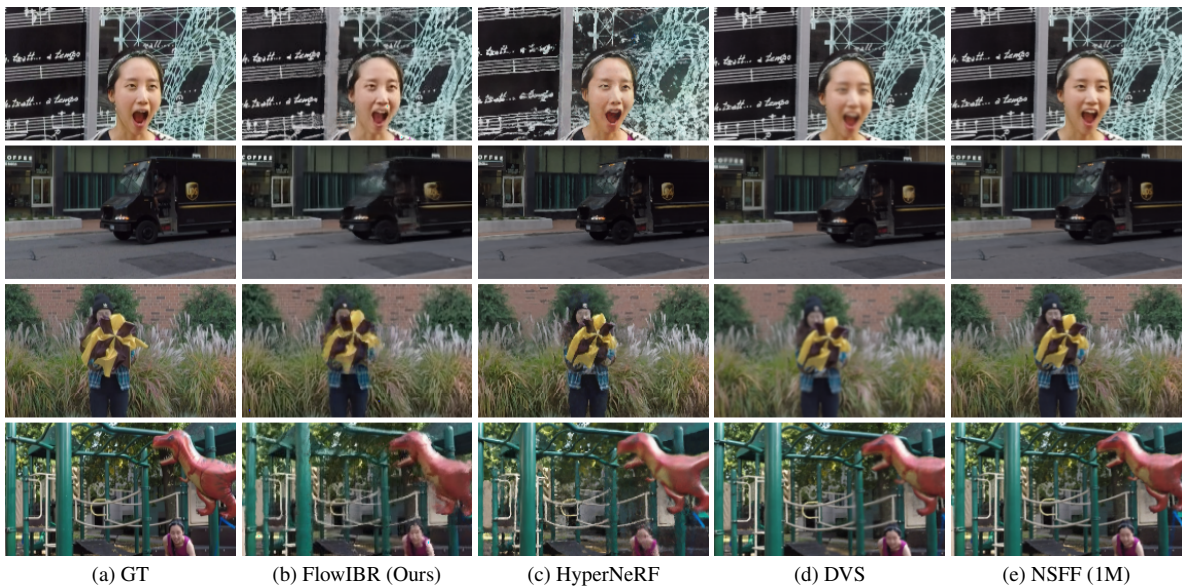}
    \caption{\textbf{
    Qualitative evaluation on Nvidia Dynamic Scenes (default) \cite{yoon2020NovelView}} Renderings for FlowIBR and the retrained methods (HyperNeRF, DVS, NSFF). We can clearly see that FlowIBR is able to synthesize novel views from previously unobserved viewpoints, with quality close to the  ground truth (GT) image and to state-of-the-art methods.}
    \label{fig:comparison}
\end{figure*}

\subsection{Experimental setup}

Optimization of the scene flow network and fine-tuning of GNT is done jointly with the Adam optimizer \cite{kingma2015AdamMethod}  on a Nvidia 3080 RTX GPU.
For the GNT backbone, we train a smaller version of the original GNT architecture with 4 transformer blocks instead of 8 and otherwise follow the training regime from Varma T. et al.~\cite{t2023AttentionAll}. 
This allows for faster rendering and the reduced memory permits a higher batch-size.
Further implementation and parameter details can be found in the supplementary materials.

As dataset for the evaluation we use the Nvidia Dynamic Scenes Dataset \cite{yoon2020NovelView}.
The eight scenes in the dataset are observed in a resolution of $1920\times1080$ by 12 static and synchronized cameras.
A moving camera is simulated by selecting one of the cameras for each timestep in a circular fashion to create the train set.
Then for each training step, one camera position is selected to sample a ray batch, while the source observations are selected from the set of remaining images.
During evaluation, observations present in the train set are kept as the source observations, while observations unseen during training are used as target images to estimate the metrics.
We utilize two versions of the dataset which differ in the temporal resolution: A \textit{default} version with 24 timesteps, and the \textit{long} version from \citeauthor{li2023DynIBaRNeural}~\cite{li2023DynIBaRNeural} with, depending on the scene, 90 to 200 timesteps.
If not explicitly mentioned, the default version is used.

Calculating the optical flow, motion masks and subsampling for the scenes takes on average 6:45min for the default and 13:51min for the long scenes.
This is similar to the set-up times of the baselines and is therefore  not included in later reported training times.

For the quantitative evaluations we report Peak Signal-to-Noise Ratio (PSNR), Structural Similarity Index Measure (SSIM) \cite{brunet2012MathematicalProperties} and Learned Perceptual Image Patch Similarity (LPIPS) \cite{zhang2018UnreasonableEffectiveness}.
In addition to estimating these metrics for the full image, we also estimate them for dynamic image regions, leveraging the motion masks from the dataset. 

\subsection{Comparative evaluation}

We benchmark FlowIBR against NSFF \cite{li2021NeuralScene}, DVS \cite{gao2021DynamicView} and HyperNeRF \cite{park2021HyperNeRFHigherdimensional}, retraining each baseline using the configurations provided by the respective authors. For DynIBaR \cite{li2023DynIBaRNeural}, which represents the current state-of-the-art in terms of rendering quality, we present the numbers reported in the original publication.
For a fair comparison in rendering time, we use images down-sampled by a factor of $f = 4$ to a resolution of $480 \times 270$ before using them for training or evaluation of the methods.

\paragraph{Qualitative results:} 
\cref{fig:comparison} presents images rendered with FlowIBR and the retrained baselines. 
FlowIBR demonstrates the capability to synthesize novel views from previously unobserved viewpoints, achieving a quality that is competitive with the state-of-the-art.
Further qualitative results are presented in the form of image sequences in the supplementary materials and as continuous renderings in the accompanying video.

\paragraph{Quantitative results:}
As the results in \cref{tab:comparison} indicate, FlowIBR performs competitively with respect to the baselines, but requires significantly shorter training time and only uses a single GPU.
In terms of rendering quality, the results are the most comparable to those of NSFF, which requires an order of magnitude more GPU hours for training.
To better understand the relative performance of the methods, we limited NSFF to a more similar training budget, causing it to underperform in comparison to FlowIBR.
The moderate rendering speed of our approach can be attributed to two primary factors.
First, we employ a generalizable rendering backbone that, although highly versatile, has not been specifically optimized for speed.
Second, our image-based rendering approach necessitates the projection of camera rays across all source observations, unlike NeRF-based methods that often render in a single forward pass.

\subsection{Ablation study}
\label{subsec:ablation}
We evaluate ablations of our method, summarized in \cref{tab:ablation}, to analyze the contribution of the different components.
The full method outperforms all ablated versions, although some ablations show high performance in individual metrics.
Most importantly, the results show that the introduction of flow compensation improves the image quality by 3.3 in PSNR for the whole image and 3.2 for the dynamic regions of the image.
We found 60k training steps to be the optimal training time as displayed in \cref{tab:time-ablation}. Shorter and longer training times lead to consistently worse metrics, especially in the dynamic parts of the image.
Replacing the permutohedral encoding with an vanilla ReLU MLP yields marginally better qualitative results, at the cost of an increased rendering time of 15.1 s/img -- more details on this can be found in the supplementary.

\begin{table}[]
    \centering
    \caption{\textbf{Method ablation}
    Ablations are obtained by selectively omitting one component at a time: (1) GNT without the scene flow, (2-3) const. subsampling $f$ instead of the coarse-to-fine subsampling, (4) optical-flow loss $\mathcal{L}_{of}$, (5) cycle loss $\mathcal{L}_{cyc}$, (6)  scene flow reg. losses $\mathcal{L}_{reg} = \mathcal{L}_{slow} + \mathcal{L}_{spat}+\mathcal{L}_{temp}$, (7) the usage of dynamic-static masks for ray sampling, and  (8) the GNT fine-tuning. 
    Metrics are averaged over the scenes \textit{Balloon1}, \textit{Truck} and \textit{Playground}.
    Best result is \textbf{bold} and second best \underline{underlined}.
    }
    \label{tab:ablation}
    \footnotesize{\begin{tabular}{w{l}{1.7cm}|w{c}{0.56cm}w{c}{0.56cm}w{c}{0.56cm}|w{c}{0.56cm}w{c}{0.56cm}w{c}{0.56cm}}
    \Xhline{3\arrayrulewidth}
    \multirow{2}{*}{Ablation}      & \multicolumn{3}{c|}{Full Image}  & \multicolumn{3}{c}{Dynamic Regions} \\
                                 &  PSNR &  SSIM & LPIPS &   PSNR&  SSIM& LPIPS \\
    \Xhline{2\arrayrulewidth}
    (1) default GNT             & 24.1 & 0.693 & 0.233 & 18.5 & 0.482 & 0.394\\
    (2) $f=4$                   & 26.0 & 0.732 & 0.157 & 19.8 & 0.589 & 0.289\\
    (3) $f=8$                   & 25.8 & 0.717 & 0.130 & 20.7 & 0.534 & 0.241\\
    (4) $\mathcal{L}_{of}$      & \textbf{26.5} & 0.781 & 0.130 & 19.8 & 0.532 & 0.281\\
    (5) $\mathcal{L}_{cyc}$     & 26.0 & 0.793 & \textbf{0.118} & \underline{21.2} & 0.610 & 0.206\\
    (6) $\mathcal{L}_{reg}$     & 26.3 & 0.795 & 0.124 & 21.0 & \underline{0.625} & \underline{0.205}\\
    (7) dyn. mask               & 26.2 & \textbf{0.803} & 0.127 & 19.7 & 0.488 & 0.275\\
    (8) fine-tune               & 25.3 & 0.782 & 0.145 & 18.5 & 0.492 & 0.258 \\
    \hline
    full                        & \underline{26.4} & \underline{0.799} & \textbf{0.118} & \textbf{21.5} & \textbf{0.631} & \textbf{0.197}\\
    \Xhline{3\arrayrulewidth}
\end{tabular}}

    \caption{\textbf{Training time ablation} Training time based parameters such as subsample frequency and learning rate decay were scaled proportionally for the ablations.
    Metrics are averaged over the scenes \textit{Balloon1}, \textit{Truck} and \textit{Playground}.}
    \vspace{-0.3cm}
    \footnotesize{
        \begin{tabular}{w{c}{0.55cm} w{c}{0.55cm} |w{c}{0.56cm}w{c}{0.56cm}w{c}{0.56cm}|w{c}{0.58cm}w{c}{0.58cm}w{c}{0.58cm}}
        \Xhline{3\arrayrulewidth}
        \multicolumn{2}{c|}{Training} & \multicolumn{3}{c|}{Full Image}  & \multicolumn{3}{c}{Dynamic Regions} \\
        steps & h  &  PSNR &  SSIM & LPIPS &   PSNR &  SSIM & LPIPS \\
        \Xhline{2\arrayrulewidth}
        30k & \textbf{0:43}   & 24.31 & 0.774 & 0.148 & 18.90 & 0.511 & 0.252\\
        60k & 1:29   & \textbf{26.42} & \textbf{0.799} & \textbf{0.118} & \textbf{21.53} & \textbf{0.631} & \textbf{0.195} \\
        90k & 2:13   & 22.57 & 0.623 & 0.211 & 19.03 & 0.483 & 0.320 \\
        \Xhline{3\arrayrulewidth}
        \end{tabular}
    }
    \label{tab:time-ablation}
    
\end{table}

\subsection{Analysis of learned scene flow}
\label{subsec:anal-scene-flow}
\begin{figure}
    \input{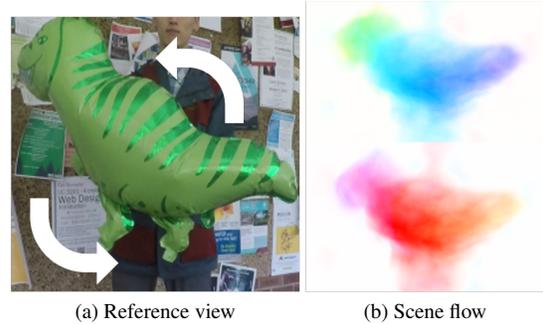}
    \caption{\textbf{Scene flow visualization} Projection of $\bm{s}_f$ (top) and $\bm{s}_b$ (bottom) onto the image plane.
    The arrows in the reference view show the true scene motion.
    Scene flow is visualized with hue indicating the direction and intensity the magnitude. Here our method correctly learned the rotating motion shown in the image.}
    \label{fig:eval-scene-flow}
\end{figure}

To analyze the learned scene flow, we visualize the forward ($\bm{s}_f$) and backward ($\bm{s}_b$) scene flow for the rays of a selected reference view in \cref{fig:eval-scene-flow}. 
We project this flow onto the image plane and color code the direction and magnitude. 
The flow is predominantly observed at pixels corresponding to the balloon, and the color-coded directions indicate a proper counter-clockwise movement, where $\bm{s}_f$ and $\bm{s}_b$ are nearly inverse to each other. 
This visualization illustrates the capability of FlowIBR in capturing correct scene dynamics.
Nevertheless, the scene flow does not exactly follow the contour of the balloon, which is one explanation for the occasional motion blur in rendered images. 

\begin{figure}
    \input{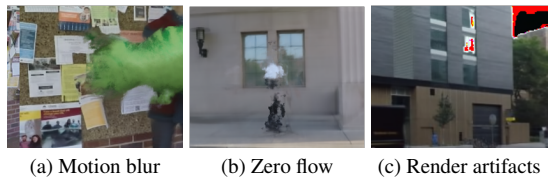}
    \caption{\textbf{Failure cases} (a) Fast moving objects can exhibit motion blur. (b) For small objects, FlowIBR occasionally fails to learn any scene flow at all, instead learning a continuous near-zero function. (c) In some cases, images contain rendering artifacts, especially for unbounded backgrounds.
    }
    \label{fig:limitations}
\end{figure}

\subsection{Limitations}
Since FlowIBR is initialized with optical flow, it inherits issues that complicate optical flow estimation, such as occlusions and undefined flow in large homogeneous areas.
Long sequences can become challenging as a singular scene flow network, restricted by its finite capacity, must capture the entire flow.
In \cref{fig:limitations} we illustrate common failure cases. 
These issues can typically be alleviated through scene-specific parameter tuning or alternative network initialization.
\section{Discussion and Conclusion}
\label{sec:conclusion}

In this paper, we presented FlowIBR, a novel view synthesis method for dynamic scenes which utilizes a pre-trained rendering method, to decrease the necessary training time.
We solve the problem of a limited number of dynamic training scenes  by employing a per-scene learned scene flow network, which is used to adjust the observations from different points in time, so they appear to be static to the rendering module.
This allows for the utilization of a readily-available rendering backbone for static scenes in the dynamic domain -- enabling shorter training times on a single consumer-grade GPU.
We would like to address the moderate rendering speed of FlowIBR in future work, by creating a specialized rendering backbone which focuses even more on fast training and rendering.

\section*{Acknowledgements}
This work was supported by the Wallenberg AI, Autonomous Systems and Software Program (WASP) funded by the Knut and Alice Wallenberg Foundation.

\newpage

{
    \small
    \bibliographystyle{ieeenat_fullname}
    \bibliography{main}
}

\clearpage
\appendix
\setcounter{section}{0}

\maketitlesupplementary

\section{Network architecture}
\begin{figure*}[!th]
    \centering
    \includegraphics[width=\textwidth]{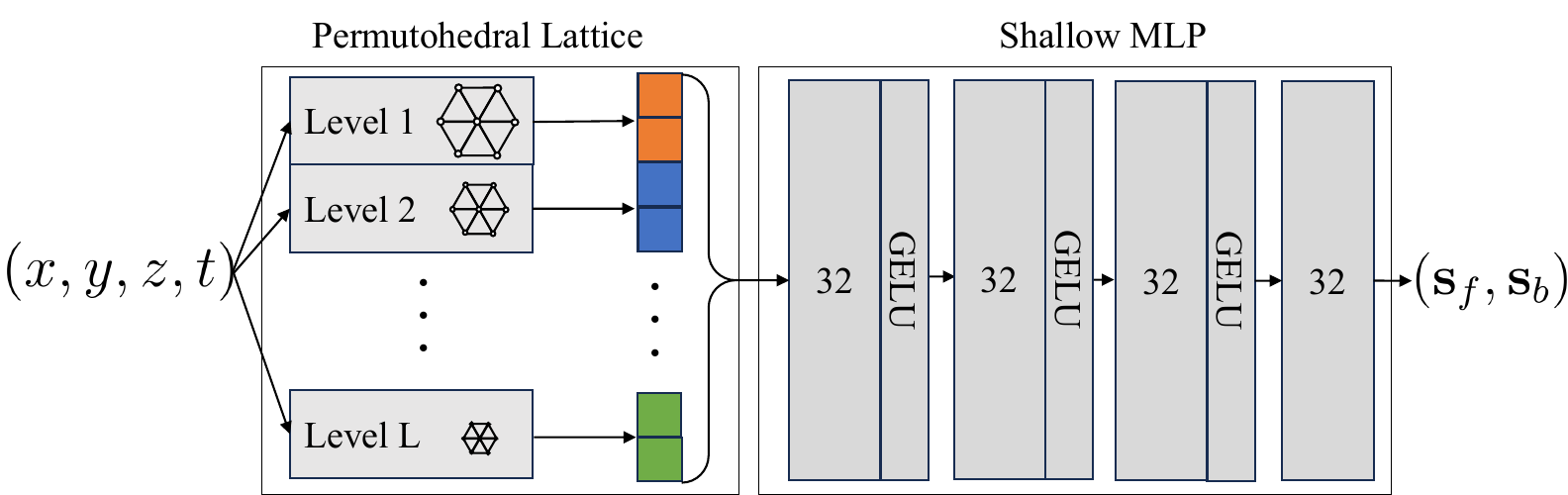}
    \caption{\textbf{Scene flow field architecture}
    For estimating the scene flow for a position and time, the 4D pose is projected into the different levels of the permutohedral lattice.
    The hashed positions of the surrounding vertices are used to extract latent vectors of length two from the respective hash-map.
    Interpolating the latent vectors of the vertices yields the final latent vector describing the local properties of the scene flow field at that level.
    The distinct latent vectors from each layer are then concatenated and inputted into a shallow 4-layer MLP with a width of 32. The MLP functions to decode the latent information to the actual scene flow.
}
    \label{fig:flow-mlp}
\end{figure*}

The scene flow network is as a combination of the permutohedral lattice from PermutoSDF \cite{rosu2023PermutoSDFFast}, which provides a latent vector for each 4D input position, and a shallow multilayer perceptron (MLP), which decodes the latent vector into the scene flow.
An overview of the network architecture is presented in \cref{fig:flow-mlp}.
The hash map of the permutohedral lattice has $L=10$ different levels, containing $T=2^{18}$ feature vectors with a dimensionality of $F=2$.

The MLP is four layers deep, with 32 nodes, and GELU activation functions \cite{hendrycks2023GaussianError} in the hidden layers.
For being applicable to general scenes, FlowIBR does not utilize normalized device coordinates (NDC) which are in the range $[0,1]$, but rather unbounded cartesian coordinates, which is the reason we use a linear activation for the final layer.
NDCs limit a method to forward facing scenes, but are common in NeRF based methods where the datasets often fulfill this criterion.
The network does not use batch normalization, dropout or any other regularization not discussed in this paper.

Images are synthesized by projecting the target rays to eight temporally close source observations, which also means projecting the rays to the eight respective times.

To show the effectiveness behind using a permutohedral lattice instead of a common architecture, we conduct an ablation by using a simple NeRF-like \cite{mildenhall2020NeRFRepresenting} ReLU MLP with depth 5 and width 128 and a sinusoidal encoding of the input.
As shown in \cref{tab:ablation_flow} the quality of the MLP and permutohedral encoding are about the same, but the rendering speed of the permutohedral encoding is substantially faster.
\vspace{-0.6cm}

\begin{table}[H]
    \centering
    \caption{\textbf{Ablation of scene flow network} Both methods were trained for 45k steps which took 2:15h for the MLP and 2:05h for the permutohedral encoding. The training parameters of the MLP were found via grid search.
    Larger MLPs yielded only marginally improvements while smaller ones decreased heavily in quality.
    Metrics are averaged over the scenes \textit{Balloon1}, \textit{Truck} and \textit{Playground}.}
    \label{tab:ablation_flow}
    \vspace{-0.3cm}
    \footnotesize{
        \begin{tabular}{l|w{c}{0.67cm}|w{c}{0.56cm}w{c}{0.56cm}w{c}{0.56cm}|w{c}{0.58cm}w{c}{0.58cm}w{c}{0.58cm}}
        \Xhline{3\arrayrulewidth}
        \multirow{2}{*}{}     & Render & \multicolumn{3}{c|}{Full Image}  & \multicolumn{3}{c}{Dynamic Regions} \\
                              & s/img  &  PSNR &  SSIM & LPIPS &   PSNR&  SSIM& LPIPS \\
        \Xhline{2\arrayrulewidth}
        MLP         & 15.1 & \textbf{26.8} & \textbf{0.805} & 0.124 & \textbf{21.8} & \textbf{0.643} & \textbf{0.193}\\
        \hline
        Permuto     & 11.8 & 26.4 & 0.799 & \textbf{0.118} & 21.5 & 0.631 & 0.197 \\
        \Xhline{3\arrayrulewidth}
        \end{tabular}
    }
\end{table}

\section{Optimization}
The network is trained over 60k steps with a batch size of 1024.
Higher batch sizes are limited by the imposed restriction of being able to train on a single GPU, lower batch sizes lead to underfitting of the scene flow.
Adam \cite{kingma2015AdamMethod} is used as the optimizer for training with $\beta_1=0.9$, $\beta_1=0.999$ and $\epsilon=10^{-8}$.
We experimented with a higher value for $\beta_1$ to increase the momentum in the running average to compensate for only having one observation per training step, but found results to be worse.
The learning rate is $l_{flow}\in[5\times10^{-3},1.0\times10^{-4}]$ for the flow network and $l_{GNT}\in[10^{-3},10^{-5}]$ for the GNT \cite{t2023AttentionAll} fine-tuning.
Every 20k steps, the learning rate is decreased by 50\%.
To continue our approach of coarse-to-fine learning of the scene flow, we follow the warm-up procedure of PermutoSDF \cite{rosu2023PermutoSDFFast} and first initialize the coarse levels and continuously include finer levels over the duration 12.5k training steps.
For pixels marked as dynamic, we reduce $\alpha_{slow}$ by $50\%$ to encourage a large scene flow in those areas. 
Additionally, we reduce the weight of the image reconstruction loss by $25\%$ for potentially dynamic pixels, to decrease blurriness in the image due to overfitting of the rendering backbone.

\section{Depth estimation}

\begin{figure*}[t]
    \centering

    \begin{subfigure}[b]{0.33\linewidth}
        \centering
        \includegraphics[width=\textwidth]{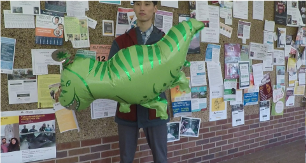}
        \caption{Ground truth}
    \end{subfigure}
    \hfill
    \begin{subfigure}[b]{0.33\linewidth}
        \centering
        \includegraphics[width=\textwidth]{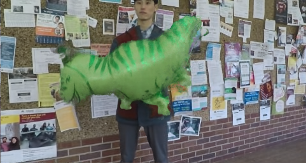}
        \caption{Rendering}
    \end{subfigure}     
    \hfill
    \begin{subfigure}[b]{0.33\linewidth}
        \centering
        \includegraphics[width=\textwidth]{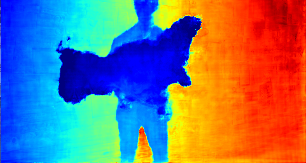}
        \caption{Depth Estimation}
    \end{subfigure}
    \caption{\textbf{Depth estimation.} After rendering the target image (b), the distances of points along the camera ray are weighted by the corresponding ray attention weights and then summed to a depth estimate (c).}
    \label{fig:depth}
\end{figure*}

Using the attention weights of the ray transformer, it is possible to infer a depth value for each pixel in the estimated image \cite{t2023AttentionAll}.
This is possible, since the ray transformer will put the most attention on the points sampled along the ray that are  contributing the most to the final pixel color, which most likely corresponds to a solid surface.
Therefore, depth can be estimated by weighting the distance of points along the ray by their respective attention weights.
An example for this is shown in \cref{fig:depth}.
In future work, this could allow for additional supervision with monocular depth estimation methods.

\section{Rendering at non-observed times}
Rendering at continuous target times $\Tilde{t} \in \mathbb{R}$ outside the intervals $\Delta t$ in which the scene has been observed, is facilitated by initiating the motion adjustment by first adjusting the continuous times to the two neighbouring observations times, with
\begin{equation}
    \Tilde{t}_b = \floor[\bigg]{\frac{\Tilde{t}}{\Delta t}}
\end{equation}
as the previous neighbour, and
\begin{equation}
    \Tilde{t}_f = \ceil[\bigg]{\frac{\Tilde{t}}{\Delta t}}
\end{equation}
as the succeeding neighbour.
With $\floor[\small]{\cdot}$ we denote the floor operator and with $\ceil[\small]{\cdot}$ the ceiling operator.

Using a scaling factor, given by:
\begin{equation}
    \delta_b = \frac{\Tilde{t}}{\Delta t} - \Tilde{t}_b,
\end{equation}
for the backward time, and respectively:
\begin{equation}
    \delta_f = \Tilde{t}_f - \frac{\Tilde{t}}{\Delta t},
\end{equation}
for the forward time to then displace the ray points from the target time $\Tilde{t}$ to the adjacent discrete observation times. 
With this, forward time adjustments can be represented as:
\begin{equation}
    \bm{p}_{\Tilde{t}\rightarrow {\Tilde{t}_f}} = \bm{p}_{\Tilde{t}} + \delta_b \mathcal{S}_f( \bm{p}_{\Tilde{t}}, \Tilde{t}_b) - \delta_f \mathcal{S}_b( \bm{p}_{\Tilde{t}}, \Tilde{t}_f)
\end{equation}
and backward time adjustments as:
\begin{equation}
    \bm{p}_{\Tilde{t}\rightarrow {\Tilde{t}_b}} = \bm{p}_{\Tilde{t}} + \delta_f \mathcal{S}_b( \bm{p}_{\Tilde{t}}, \Tilde{t}_f) - \delta_b \mathcal{S}_f( \bm{p}_{\Tilde{t}}, \Tilde{t}_b) \text{ .}
\end{equation}
Following this initial step, the motion adjustment can continue as described for discrete times.

\section{Images sequences}

\begin{figure*}[b]
    \centering
    \input{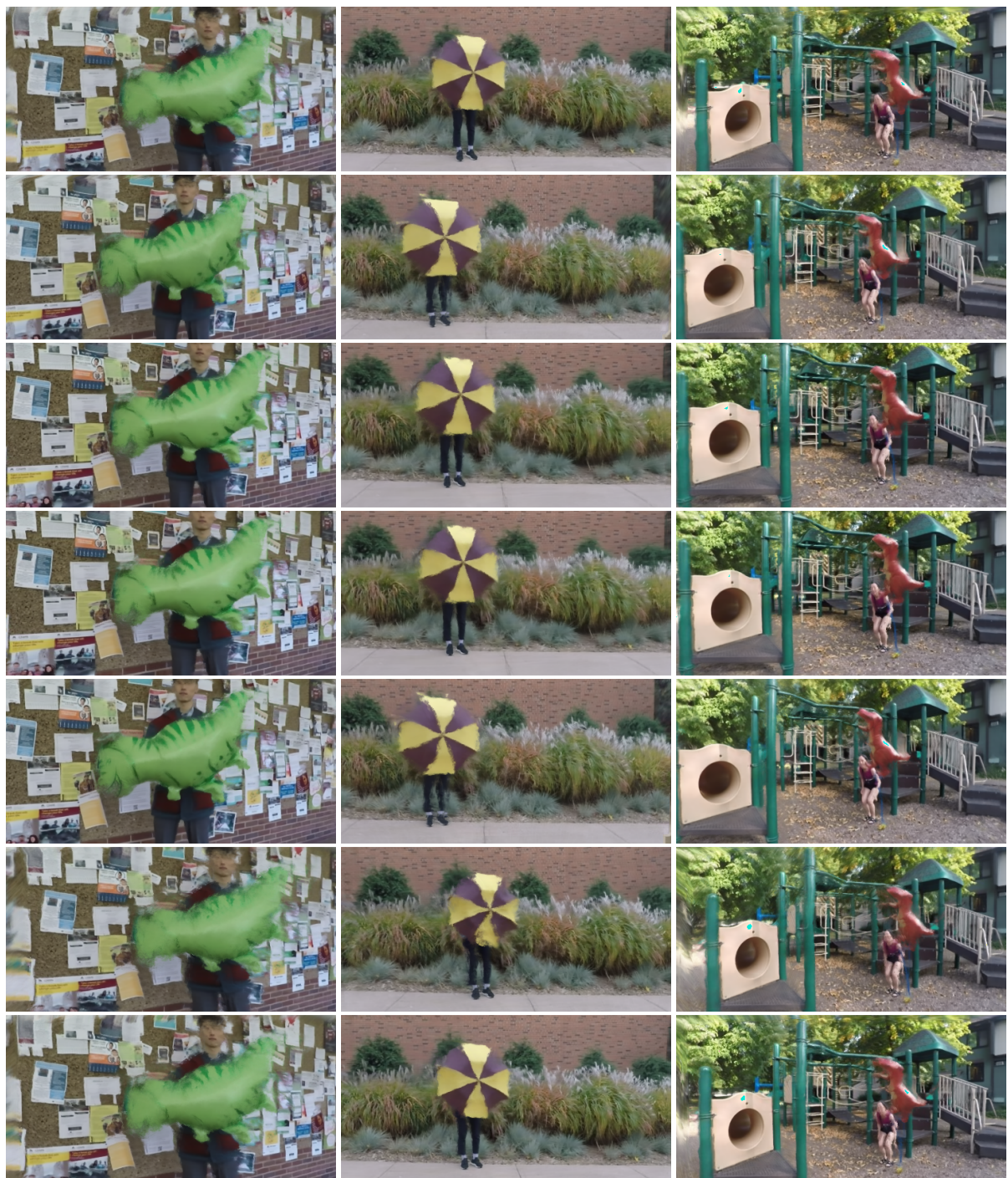}
    \caption{\textbf{Image sequences: Bullet time} 
    Three image sequences of bullet time renderings, where the viewpoint is continuously changed while the time stays constant.
    Viewpoints are on an elliptical trajectory along the center of the camera rig.}
    \label{fig:sequences-bt}
\end{figure*}

In \cref{fig:sequences-bt} we present image sequences of view interpolation for three different scenes.
Overall, the method is able to consistently interpolate between the different target viewpoints without the sudden appearance of artifacts or flickering of the scene content.
Nevertheless, the further the target viewpoint gets away from the nearest source observation, the more the dynamic content starts to blur because of the missing information from that perspective.


\section{Societal Impact}
We anticipate several potential impacts of our proposed method, and similar methods, on society in the future.
Primarily, we present a method to decrease the necessary training time for novel view synthesis methods for dynamic scenes and allow training on a single consumer-grade GPU which  has the potential to democratize research capabilities, alleviating the dependency on specialized hardware.
While our approach relies on the accumulated knowledge from pre-training, it is imperative to note that any biases present in the used datasets might propagate through our system.
However, as the essence of our research is to synthesize views that are as faithful to the ground truth as possible, we are actively addressing this challenge to ensure accuracy and integrity of our results.

\end{document}